%% file: main.tex
\documentclass[sigconf]{acmart}

\usepackage{amsfonts}
\usepackage{comment}
\usepackage{amsmath}
\usepackage{subcaption} 
\usepackage[T1]{fontenc}
\usepackage{enumitem}
\usepackage{booktabs}
\usepackage{makecell}
\usepackage{graphicx}
\usepackage{mathtools}
\usepackage{caption} 

\DeclareUnicodeCharacter{2212}{-}

\usepackage[ruled]{algorithm2e}

\AtBeginDocument{%
  \providecommand\BibTeX{{%
    \normalfont B\kern-0.5em{\scshape i\kern-0.25em b}\kern-0.8em\TeX}}}

\copyrightyear{2021}
\acmYear{2021}
\setcopyright{acmlicensed}\acmConference[KDD '21]{Proceedings of the 27th ACM SIGKDD Conference on Knowledge Discovery and Data Mining}{August 14--18, 2021}{Virtual Event, Singapore}
\acmBooktitle{Proceedings of the 27th ACM SIGKDD Conference on Knowledge Discovery and Data Mining (KDD '21), August 14--18, 2021, Virtual Event, Singapore}
\acmPrice{15.00}
\acmDOI{10.1145/3447548.3467409}
\acmISBN{978-1-4503-8332-5/21/08}

\begin{document}

\newcommand{\Require}{\textbf{Input:}\,}
\newcommand{\Ensure}{\textbf{Output:}\,}
\newcommand{\overbar}[1]{\mkern 1.5mu\overline{\mkern-1.5mu#1\mkern-1.5mu}\mkern 1.5mu}
\newcommand{\x}{\boldsymbol{x}}
\newcommand{\y}{\boldsymbol{y}}
\newcommand{\z}{\boldsymbol{z}}
\newcommand{\e}{\boldsymbol{e}}
\newcommand{\model}{CrowdInG}

\fancyhead{}

\title{Improve Learning from Crowds via Generative Augmentation}


\author{Zhendong Chu, Hongning Wang}
\affiliation{%
    \institution{Department of Computer Science}
  \institution{University of Virginia}
  \city{Charlottesville}
  \state{VA 22903}
  \country{USA}
  }
\email{{zc9uy, hw5x}@virginia.edu}

\renewcommand{\shortauthors}{Chu and Wang}

\begin{abstract}
Crowdsourcing provides an efficient label collection schema for supervised machine learning. However, to control annotation cost, each instance in the crowdsourced data is typically annotated by a small number of annotators. This creates a \emph{sparsity} issue and limits the quality of machine learning models trained on such data. In this paper, we study how to handle sparsity in crowdsourced data using data augmentation. Specifically, we propose to directly learn a classifier by augmenting the raw sparse annotations. We implement two principles of high-quality augmentation using Generative Adversarial Networks: 1) the generated annotations should follow the distribution of authentic ones, which is measured by a discriminator; 2) the generated annotations should have high mutual information with the ground-truth labels, which is measured by an auxiliary network. Extensive experiments and comparisons against an array of state-of-the-art learning from crowds methods on three real-world datasets proved the effectiveness of our data augmentation framework. It shows the potential of our algorithm for low-budget crowdsourcing in general.
\end{abstract}




\maketitle

\input{introduction}

\input{related_works}
\input{method}

\input{optimization}
\input{experiments}

\section{Conclusions \& Future works}
Data sparsity poses a serious challenge to current learning from crowds solutions. We present a data augmentation solution using generative adversarial networks to handle the issue. We proposed two important principles in generating high-quality annotations: 1) the generated annotations should follow the distribution of authentic ones; and 2) the generated annotations should have high mutual information with the ground-truth labels. We implemented these principles in our discriminative model design. Extensive experiment results demonstrated the effectiveness of our data augmentation solution in improving the performance of the classifier learned from crowds, and it sheds light on our solution's potential in low-budget crowdsourcing in general. 

Our exploration also opens up a series interesting future directions. As our generative module captures annotator- and instance-specific confusions, it can be used for annotator education \cite{singla2014near}, e.g., inform individual annotators about their potential confusions. Our solution can also be used for interactive labeling with annotators \cite{yan2011active}, e.g., only acquire annotations on which our generative module currently has a low confidence. Also, the instance-level confusion modeling can better support fine-grained task assignment \cite{deng2013fine}, e.g., gather senior annotators for specific tasks.

\section*{Acknowledgement}
This work was partially supported by the National Science Foundation under award NSF IIS-1718216 and NSF IIS-1553568, and the Department of Energy under the award DoE-EE0008227.

\bibliographystyle{ACM-Reference-Format}
\bibliography{sample-base}

\end{document}

%% file: introduction.tex
\section{Introduction}
Modern machine learning systems are data hungry, especially for labeled data, which unfortunately is expensive to acquire at scale. Crowdsourcing provides a label collection schema that is both cost- and time-efficient \cite{buecheler2010crowdsourcing}. It spurs the growing research efforts in directly learning a classifier with only crowdsourced annotations, aka the \emph{learning from crowds} problem.

In practice, to minimize annotation cost, the instances in crowdsourced data are typically labeled by a small number of annotators; and each annotator will only be assigned to a few instances. This introduces serious sparsity in crowdsourced data. We looked into two widely-used public crowdsourced datasets for multi-class classification, one for image labeling (referred to as LabelMe \cite{russell2008labelme, rodrigues2018deep}) and one for music genre classification (referred to as Music \cite{rodrigues2014gaussian}). On the LabelMe dataset, each instance is only labeled by 2.5 annotators on average (out of 59 annotators), while 88\% annotators provide less than 100 annotations (out of 1,000 instances). On the Music dataset, each instance is labeled by 4.2 annotators on average (out of 44 annotators), while 87.5\% annotators provide less than 100 annotations (out of 700 instances). 
Such severe sparsity hinders the utility of  crowdsourced labels. On the instance side, annotations provided by non-experts are noisy, which are expected to be improved by redundant annotations. But subject to the budget constraint, redundancy is also to be minimized. This conflict directly limits the quality of crowdsourced labels. On the annotator side, most existing crowdsourcing algorithms model annotator-specific confusions, which are used for label aggregation \cite{dawid1979maximum}, task assignment \cite{deng2013fine, li2016crowdsourced} and annotator education \cite{singla2014near}. But due to the limited observations per annotator, such modeling can hardly be inaccurate,  and thus various approximations (e.g., strong independence assumptions \cite{dawid1979maximum}) have to be devised. 


A straightforward solution to address annotation sparsity is to recruit more annotators or increase their assignments, at the cost of an increasing budget. This however is against the goal of crowdsourcing, i.e., to collect labeled data at a low cost. We approach the problem from a different perspective: we perform data augmentation using generative models to fill in the missing annotations. Instead of collecting more real annotations, we generate annotations by modeling the annotation distribution on instances and annotators. Given our end goal is to obtain an accurate classifier, the key is to figure out \emph{what annotations best help the classifier's training}. We propose two important criteria. 
First, the generated annotations should follow the distribution of authentic ones, such that they will be consistent with the label confusion patterns observed in the original annotations. 
Second, the generated annotations should well align with the ground-truth labels, e.g., with high mutual information \cite{xu2019l_dmi, harutyunyan2020improving}, so that they will be informative about ground-truth labels to the classifier. 


We realize our criteria for annotation augmentation in crowdsourced data using Generative Adversarial Networks (GAN) \cite{goodfellow2014generative}. The end product of our solution is a classifier, which predicts the label of a given instance.
We set a discriminative model to judge whether an annotation is authentic or generated. Meanwhile, a generative model aims to generate annotations following the distribution of authentic annotations under the guidance of the discriminative model. 
On a given instance, the generator takes the classifier's output and the annotator and instance features as input to generate the corresponding annotation. 
To ensure the informativeness of generated annotations, we maximize the mutual information between the classifier's predicted label and the generated annotation on each instance \cite{chen2016infogan}. A two-step training strategy is proposed to avoid model collapse. We name our framework as \emph{CrowdInG} - learning with \underline{\textbf{Crowd}}sourced data through \underline{\textbf{In}}formative \underline{\textbf{G}}enerative augmentation. Extensive experiments on three real-world datasets demonstrated the feasibility of data augmentation for the problem of learning from crowds. Our solution outperformed a set of state-of-the-art crowdsourcing algorithms; and its advantage becomes especially evident with extremely sparse annotations. It provides a new opportunity for low-budget  crowdsourcing in general.

%% file: related_works.tex
\section{Related works}
Our work studies the learning from crowds problem. \citet{raykar2010learning} employed an EM algorithm to jointly estimate the expertise of different annotators and a logistic regression classifier on crowdsourced data. They followed the well-known Dawid and Skene (DS) model \cite{dawid1979maximum} to model the observed annotations. \citet{albarqouni2016aggnet} extended this solution by replacing the logistic classifier with a deep neural network classifier. \citet{rodrigues2018deep} further extended the solution by replacing the confusion matrix in the DS model with a neural network to model annotators' expertise, and trained the model in an end-to-end manner. \citet{guan2018said} used a neural classifier to model each annotator, and aggregated the predictions from the classifiers by a weighted majority vote. \citet{cao2018maxmig} proposed an information-theoretical deep learning solution to handle the correlated mistakes across annotators. However, all the mentioned solutions only use the observed annotations, such that their practical performance is limited by the sparsity of annotations. 

Another research line focuses on modeling the annotators. \citet{whitehill2009whose} proposed a probabilistic model which considers both annotator accuracy and instance difficulty. \citet{rodrigues2014gaussian} modeled the annotation process by a Gaussian process. \citet{imamura2018analysis} and \citet{venanzi2014community} extended the DS model by sharing the confusion matrices among similar annotators to improve annotator modeling with limited observations. Confusions of annotators with few annotations are hard to be modeled accurately, and \citet{kamar2015identifying} proposed to address the issue with a shared global confusion matrix. \citet{chu2020learning} also set a global confusion matrix, which is used to capture the common confusions beyond individual ones. However, the success of the aforementioned models relies on the assumed structures among annotators or annotations. Such strong assumptions are needed, because the sparsity in the annotations does not support more complicated models. But they also restrict the modeling of crowdsourced data, e.g., introducing bias in the learnt model. We lift such restrictions by directly generating annotations, such that our modeling of crowdsourced data even does not make any class- or annotator-dependent assumptions. 

Benefiting from their powerful modeling capabilities, deep generative models have been popularly used for data augmentation purposes. Most efforts have been spent on problems in a continuous space, such as image and video generations. Semi-supervised GAN \cite{odena2016semi, springenberg2015unsupervised, antoniou2017data} augments training data by generating new instances from labeled ones.  \citet{chae2018cfgan} employed GAN to address the data sparsity in content recommendation, with their proposed real-value, vector-wise recommendation model training. Recently, GAN has also been adopted in data augmentation for discrete problems. \citet{wang2019enhancing} designed a two-step solution to perform GAN training for collaborative filtering. \citet{wang2018graphgan} unified generative and discriminative graph neural networks in a GAN framework to enhance the graph representation learning. \citet{irissappane2020leveraging} reduced the needed labeled data to fine-tune the BERT-like text classification models via GAN-generated examples. 

%% file: method.tex
\begin{figure*}
    \centering
    \includegraphics[width=15cm]{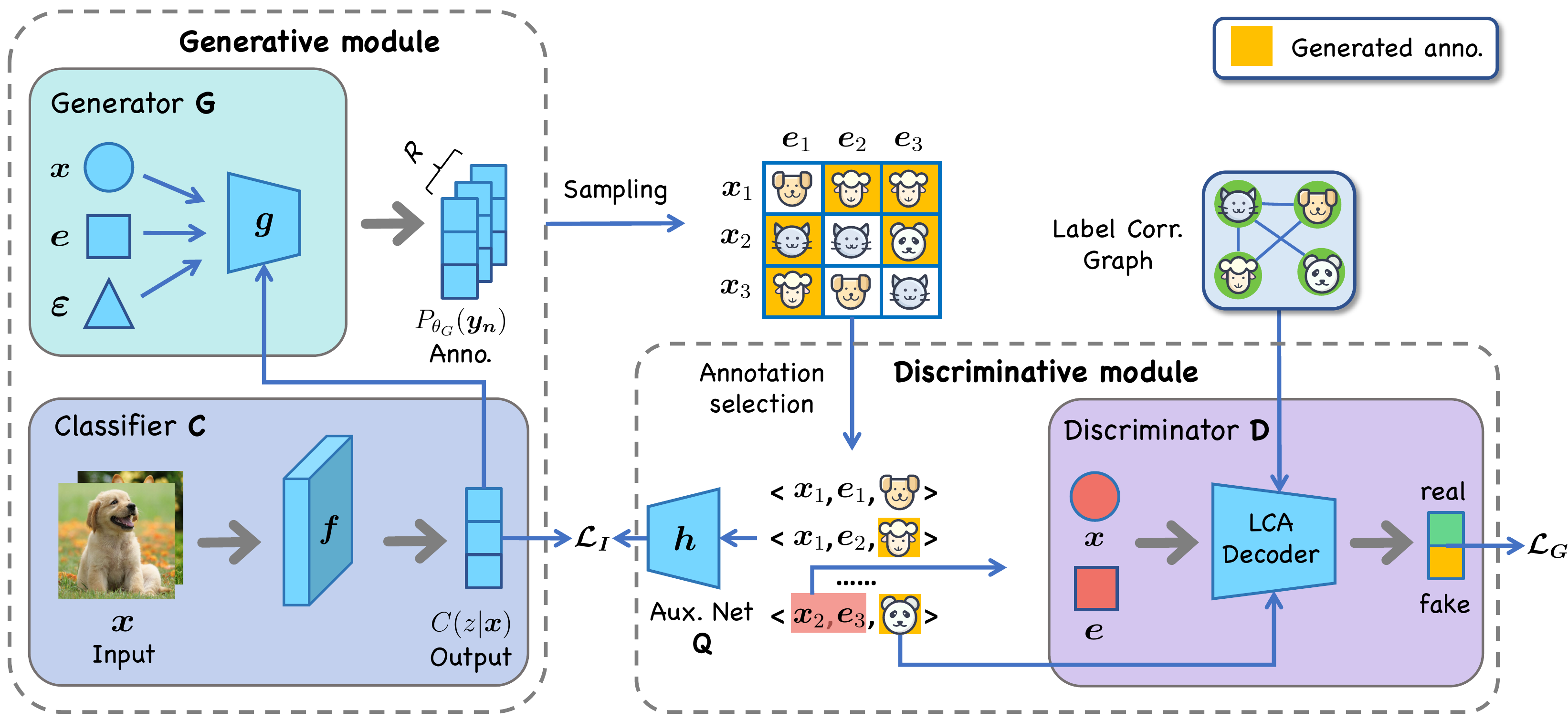}
    \caption{Overview of CrowdInG framework. We first sample annotations from annotation distributions provided by the generator. The discriminator and the auxiliary network are trained on the selected annotations. Then, the classifier is first fixed and the generator is updated according to $\mathcal{L}_G$ and $\mathcal{L}_I$. The generator is fixed and the classifier is updated according to $\mathcal{L}_G$.}
    \label{fig:overview}
    \vspace{-0.5em}
\end{figure*}

\section{Methodology}
In this section, we begin our discussion with the background techniques of our augmentation framework, including GAN and InfoGAN. Then we present our CrowdInG solution and discuss its detailed design of each component. Finally, we describe our two-step training strategy for the proposed framework.

\subsection{Background}
\subsubsection{Generative Adversarial Networks} \citet{goodfellow2014generative} introduced the GAN framework for training deep generative models as a \emph{minimax} game, whose goal is to learn a generative distribution $P_G(x)$ that aligns with the real data distribution $P_{\text{true}}(x)$. The generative distribution is imposed by a generative model $G$, which transforms a noise variable $\boldsymbol{\varepsilon}\sim P_{noise}(\boldsymbol{\varepsilon})$ into a sample $G(\boldsymbol{\varepsilon})$. A discriminative model $D$ is set to distinguish between the authentic and generated samples. The generator $G$ is trained by playing against the discriminator $D$. Formally, $G$ and $D$ play the following two-player minimax game with value function $V(G, D)$:
\begin{align}
        \underset{G}{\text{min}}\,\underset{D}{\text{max}}\,V(G, D) =& \mathbb{E}_{x\sim P_{\text{true}}}[\phi\big(D(x)\big)] + \mathbb{E}_{\boldsymbol{\varepsilon}\sim P_{\text{noise}}}[\phi\big (1-D(G(\boldsymbol{\varepsilon}))\big )], \nonumber
\label{eq:value}
\end{align}
where $\phi$ is a function of choice and $\text{log}(\cdot)$ is typically the choice. The optimal parameters of the generator and the discriminator can be learned by alternately maximizing and minimizing the value function $V(G, D)$. In this paper, we adopt this idea to model the annotation distribution: a generator is used to generate annotations on specific instances and annotators; and a discriminator is set to distinguish the authentic annotations from the generated ones. 

\subsubsection{Information Maximizing Generative Adversarial Networks} \citet{chen2016infogan} extended GAN with an information-theoretic loss to learn disentangled representations for improved data generation. Aside from the value function $V(G, D)$, InfoGAN also maximizes the mutual information between a small subset of latent variables (referred to as latent code $z$) and the generated data. The generator takes both random noise $\boldsymbol{\varepsilon}$ and latent code $z$ as input, where the latent code is expected to capture the salient structure in the data distribution. The minimax game then turns into an information-regularized form,
\begin{align*}
        \underset{G}{\text{min}}\,\underset{D}{\text{max}}\,V_I(G, D) = V(G, D) - \lambda I(z; G( \boldsymbol{\varepsilon}, z)),
\end{align*}
where $I(x;y)$ is the mutual information between random variables $x$ and $y$, and $\lambda$ is the regularization coefficient. 

\subsection{The CrowdInG framework}
Let $\mathcal{S}=\{\x_n, \y_n\}_{n=1}^N$ denote a set of $N$ instances labeled by $R$ annotators out of $|\mathcal{C}|$ possible classes. We define $\x_n \in \mathbb{R}^d$ as the feature vector of the $n$-th instance and $y_n^r \in \mathcal{C}$ as its annotation provide by the $r$-th annotator. $\y_n$ is thus the annotation vector (with missing values) from $R$ annotators for the $n$-th instance. When available, the feature vector of the $r$-th annotator is denoted as $\e_r$; otherwise, we use a one-hot vector to represent an annotator. 
Each instance is associated with an unobserved ground-truth label $z\in \mathcal{C}$. 
The goal of learning from crowds is to obtain a classifier $C(z|\x)$ that is directly estimated from $\mathcal{S}$. 

The framework of CrowdInG is depicted in Figure \ref{fig:overview}. It consists of two main components: 1) a generative module, including a classifier and a generator;
and 2) a discriminative module, including a discriminator and an auxiliary network.
In the generative module, the classifier first takes an instance $\x_n$ as input and outputs its predicted label distribution $P_{\theta_C}(z_n|\x_n)$. For simplicity, we collectively denote classifier's output for an instance $\x_n$ as $\hat{\z}_n$. And then the generator takes the instance $\x_n$, annotator $\e_r$, the classifier's output $\hat{\z}_n$, together with a random noise vector $\boldsymbol{\varepsilon}$, to generate the corresponding annotation distribution $P_{\theta_G}(y_n^r|\x_n, \e_r, \hat{\z}_n, \boldsymbol{\varepsilon})$. 
The discriminative module is designed based on our criteria of high-quality annotations to evaluate the generations. On one hand, the discriminative module uses a discriminator to differentiate whether the annotation triplet ($\x_n, \e_r, y_n^r$) is authentic or generated. On the other hand, the discriminative module penalizes the generation based on the mutual information between the generated annotation and classifier's output measured by an auxiliary network. Following the idea of InfoGAN, we treat the classifier's output $\hat{\z}$ as the latent code in our annotation generation. And the auxiliary network measures the mutual information between $\hat{\z}$ and $y$. The two modules play a minimax game in CrowdInG. A better classifier is expected as the discriminative module faces more difficulties in recognizing the generated annotations during training. 

\subsubsection{Generative module} The output of the generative module is an annotation distribution for a given annotator-instance pair $(\x_n, \e_r)$. Sampling is applied to obtain the final annotations. As shown in Figure \ref{fig:overview}, this is a two-step procedure.
First, the classifier $C(z_n|\x_n;\theta_C)$ predicts the label of a given instance $\x_n$ by
\begin{equation*}
    P_{\theta_C}(z_n=c | \x_n) \propto \exp[ f(\x_n, z_n=c)],
\end{equation*}
where $f(\cdot)$ is a learnable scoring function chosen according to the specific classification tasks. Then the generator $G$ takes the classifier's output $\hat{\z}$ as input to predict the underlying annotation distribution for the given annotator-instance pair. Moving beyond the classical class-dependent annotation confusion assumption \cite{dawid1979maximum, rodrigues2018deep}, we impose a much more relaxed generative process about the annotations. We consider the confusions can be caused by instance difficulty, or annotator expertise, or true labels of the instances (e.g., different annotation difficulty in different label categories), or even some random noise. To realize the idea, we provide the feature vector $\x_n$ of the instance, the annotator $\e_r$ and the classifier's output $\hat{\z}_n$ to the generator as input, and the corresponding annotation distribution is modeled as,
\begin{equation}
    P_{\theta_G}(y_n^r=c|\x_n, \e_r, \boldsymbol{\varepsilon}, \hat{\z}_n) \propto \exp [g(y_n^r=c, \x_n, \e_r, \boldsymbol{\varepsilon}, \boldsymbol{\hat{z}}_n)],
    \label{eq:gen_prob}
\end{equation}
where $\boldsymbol{\varepsilon} \sim \mathcal{N}(0, 1)$ is a random noise vector, $g(\cdot)$ is a learnable scoring function implemented via a neural network. The generated annotations are sampled from the resulting distribution $P_{\theta_G}$. To simplify our notations, we use $G(\x_n, \e_r, \boldsymbol{\varepsilon}, \hat{\z}_n)$ to represent the predicted annotation distribution; and when no ambiguity is invoked, we denote $G(y_n^r)$ as its $c$-th entry when $y_n^r=c$. Thanks to our data augmentation framework, we can afford a more flexible modeling of the annotation noise, e.g., dropping the hard independence assumptions made in previous works \cite{rodrigues2018deep, dawid1979maximum}. This in turn helps us boost the quality of generated annotations.

\subsubsection{Discriminative module} We realize our principles of high-quality annotations in the discriminative module. First, the discriminator $D$ aims to differentiate whether an annotation $y^r_n$ is authentic from annotator $\e_r$ to instance $\x_n$, i.e., $D(y^r_n| \x_n, \e_r; \theta_D)$ predicts the probability of annotation $y^r_n$  being authentic. In a crowdsourcing task, an annotator might confuse a ground-truth label with several classes, such that all of the confused classes could be independently authentic. For example, if an annotator always confuses ``birds'' with ``airplanes'' in low resolution images, his/her annotations might be random between these two categories. And thus both types of annotations should be considered as valid, as there is no way to tell which annotation is ``correct'' only based on the observations of his/her annotations. As a result, we realize the discriminator as a multi-label classifier, which takes an annotation triplet ($\x_n, \e_r, y_n^r$) as input and calculates the discriminative score by a bilinear model,
\begin{align}
\label{eq_discriminator}
    D(y_n^r = c| \x_n, \e_r; \theta_D) = \sigma(\boldsymbol{u}_r^\top\boldsymbol{M}_c\boldsymbol{v}_n),
\end{align}
\begin{equation*}
        \boldsymbol{u}_r = \boldsymbol{W}_u\e_r + b_u, \boldsymbol{v}_n = \boldsymbol{W}_v\x_n + b_v, \nonumber
\end{equation*}
where $\sigma(\cdot)$ is the sigmoid function, $\boldsymbol{M}_c$ is the weight matrix for class $c$, ($\boldsymbol{W}_v, b_v$) and ($\boldsymbol{W}_u, b_u$) are weight matrices and bias terms for annotator and instance embedding layers. For simplicity, we denote $D(y_n^r)$ as the discriminator's output on annotation $y_n^r$. 

However, Eq \eqref{eq_discriminator} does not consider the correlation among different classes in the annotations, as it still evaluates each possible label independently. The situation becomes even worse with sparse observations in individual annotators. For example, when an annotator confuses ``bird'' with ``airplanes'', the discriminator might decide the label of ``bird'' is more authentic for this annotator, simply because this category appears more often in the annotator's observed annotations. To capture such ``\emph{equally plausible}'' annotations, we equip the discriminator with additional label correlation information \cite{lanchantin2019neural}. Specifically, we use a graph convolution network (GCN) \cite{kipf2016semi} to model label correlation. Two labels are more likely to be correlated if they are provided to the same instance (by different annotators) in the authentic annotations. We calculate the frequency of label co-occurrence in the observed annotations to construct the adjacency matrix $\boldsymbol{A}$.
Then we extend the weight matrix $\boldsymbol{M}_c$ in Eq \eqref{eq_discriminator} by $\hat{\boldsymbol{M}}_c = \hat{\boldsymbol{D}}^{-\frac{1}{2}}\hat{\boldsymbol{A}}\hat{\boldsymbol{D}}^{-\frac{1}{2}}\boldsymbol{M}_c\boldsymbol{W}$, with $\hat{\boldsymbol{A}} = \boldsymbol{A} + \boldsymbol{I}$ where $\boldsymbol{I}$ is the identity matrix, $\hat{D}$ is the diagonal node degree matrix of $\hat{\boldsymbol{A}}$. We name this component as the label correlation aggregation (LCA) decoder. We also enforce sparsity on the discriminator by applying L2 norm to its outputs. 

To realize our second criterion, an auxiliary network $Q$ is used to measure the mutual information between the classifier's prediction $\hat \z_n$ and the generated annotation $y^r_n$ on instance $\x_n$. To simplify our notations in the subsequent discussions, we denote $G(\x_n, \e_r, \boldsymbol{\varepsilon}, \hat{\z}_n)$ as $G(\boldsymbol{\varepsilon}, \hat{\z})$ to represent the annotation distribution predicted on pair $(\x_n, \e_r)$. As our generator design is very flexible to model complex confusions, it however becomes useless for classifier training if the learnt confusions are independent from the classifier's outputs. For example, if the generator learnt to generate a particular annotation only by the annotator's features (e.g., the most frequently observed label in this annotator), such a generation contributes no information to classifier training. We propose to penalize such generations by maximizing the mutual information between classifier's output and the generated annotations for an instance, i.e., $I(\hat{\z}; G(\boldsymbol{\varepsilon}, \hat{\z}))$. 

In practice, mutual information is generally difficult to optimize, because it requires the knowledge of posterior $P(\hat{z}|y)$. We follow the design in \cite{chen2016infogan} to maximize the variational lower bound of $I(\hat{\z}; G(\boldsymbol{\varepsilon}, \hat{\z}))$ by utilizing an auxiliary distribution $P_Q$ to approximate $P(\hat{z}|y)$:
\begin{align}
\label{eq:info_lb}
    \mathcal{L_I}(G, Q) &= \mathbb{E}_{\hat{z}\sim P(\hat{\z}), y\sim G(\boldsymbol{\varepsilon}, \hat{\z})}[\log P_Q(\hat{z}|y)] + H(\hat{z}) \\
    &\leq I(\hat{\z}; G(\boldsymbol{\varepsilon}, \hat{\z})).\nonumber
\end{align}

We refer to $\mathcal{L}_I$ as the information loss, which can be viewed as an information-theoretical regularization to the original minimax game. The auxiliary distribution $P_Q(\hat{z}|y)$ is parameterized by the auxiliary network $Q$. In our implementation, we devise a two-step training strategy for the entire pipeline (details in Section \ref{sec:two-step}), where we fix the classifier when updating the generator. As a result, $H(\hat{z})$ becomes a constant when updating the generator by Eq \eqref{eq:info_lb}. Since the posterior $P(\hat{z}|y)$ can be different when the annotations are given by different annotators on different instances, we also provide the instance and annotator features to the auxiliary network, 
\begin{equation*}
    P_{\theta_Q}(\hat{z}_n=c | \x_n, \e_r, y_n^r) \propto \exp[h(\hat{z}_n=c, \x_n, \e_r, y_n^r)],
\end{equation*}
where $h(\cdot)$ is a learnable scoring function. To reduce model complexity, we reuse the annotator and instance encoding layers from the discriminator here. The class-related weight matrix $\hat{\boldsymbol{M}}_c$ is flatten and transformed to a low-dimension embedding $\boldsymbol{m}_c$ by an embedding layer for each annotation type $y_n^r=c$.


Putting the generative and discriminative modules together, we formalize the value function of our minimax game for learning from crowds in \model{} as,
\begin{align}
\label{eq_obj}
&\underset{C, G, Q }{\text{min}}\,\underset{D}{\text{max}}\,V_\text{CrowdInG}(C, G, D, Q) =V(C, G, D) - \lambda \mathcal{L}_I(G, Q)\\
&V(C, G, D) = \mathbb{E}_{y\sim P_{\text{true}}}[\log\big(D(y)\big)] + \mathbb{E}_{\boldsymbol{\varepsilon}\sim P_{\text{noise}}, y\sim P_{G(\boldsymbol{\varepsilon}, \hat{\z})}}[\log\big (1-D(y)\big )], \nonumber
\end{align}
where $\lambda$ is a hyper-parameter to control the regularization. The value function is maximized by updating the discriminator to improve its ability in differentiating the authentic annotations from the generated ones, and minimized by updating the classifier, generator and the auxiliary network to generate more high-quality annotations.

%% file: optimization.tex
\subsection{Model optimization}
In this section, we introduce the training strategy for \model, which cannot be simply performed via vanilla end-to-end training. First, the number of unobserved annotator-instance pairs is much larger than the observed ones. Blindly using all the generated annotations overwhelms the training of our discriminative module, and simply leads to trivial solutions (e.g., classifying all annotations as generated). As our solution, we present an entropy-based annotation selection strategy to select informative annotations for discriminative module update. Second, due to the required sampling procedure when generating the annotations, there are non-differentiable steps in our generative module. We resort to an effective counterfactual risk minimization (CRM) method to address the difficulty. Finally, the classifier and the generator in the generative module might change dramatically to fit the complex training signals, which can easily cause model collapse. We propose a two-step training strategy to prevent it in practice. 

\subsubsection{Entropy-based annotation selection} We borrow the idea from active learning \cite{settles2009active}: \emph{the discriminator should learn to distinguish the most difficult annotations}. A generated annotation is more difficult for the discriminator if the generator is more confident about it. Formally, the selection strategy is designed as,
\begin{equation*}
    P(y_n^r) \propto \frac{1}{H(G(\x_n, \e_r, \boldsymbol{\varepsilon}, \hat{\z}_n))}
    \label{eq:dis_anno_sel},
\end{equation*}
where $H(G(\x_n, \e_r, \boldsymbol{\varepsilon}, \hat{\z}_n))$ is the entropy of the annotation distribution. To reduce training bias caused by annotation sparsity in individual annotators, we sample the same number of generated annotations as the authentic ones in each annotator. As a by-product, our instance selection also greatly reduces the size of training data for the discriminative module. It makes discriminator training a lot more efficient. To fully utilize the power of discriminative module, we use all generated annotations for the generator updating. 

\subsubsection{Gradient-based optimization} The gradient for the discriminator and the auxiliary network is easy to compute by calculating the derivative on trainable parameters. However, due to the required sampling steps for generating specific annotations, there are non-differentiable steps in the generative module. Previous works \cite{wang2018graphgan, wang2019enhancing} use Gumbel-softmax trick or policy gradient to handle the non-differentiable functions. However, once the generator is updated, we need to re-sample the annotations and evaluate them again using the discriminative module, which is time-consuming. To accelerate our model training, we perform batch learning from logged bandit feedback \cite{joachims2018deep, swaminathan2015batch}. In each epoch, we treat the generative module from the last epoch as the logging policy $G_0$, and sample annotations from it. Because the discriminator only evaluates the sampled annotations from the (last) generative module, rather than the entire distribution of annotations predicted by the module, training signals received on the generative module side are in the form of logged bandit feedback. 

When updating the generator, the training signals are from both the discriminator $\mathcal{L}_G = \log\big( 1-D(y)\big )$ and the information loss $-\lambda \mathcal{L}_I$. We collectively denote them as loss $\delta=\mathcal{L}_G-\lambda \mathcal{L}_I$. In each epoch, we update the generator $G_{\theta_G}$ as follows, 
\begin{align}
    \theta_G = \underset{\theta_G}{\text{argmin}}\frac{1}{NR}\sum_{n=1}^N\sum_{r=1}^R\frac{\big(\delta(y_n^r)-\mu\big)G_{\theta_G}(y_n^r)}{G_0(y_n^r)},
    \label{eq:gen_ips}
\end{align}
where $\mu$ is a Lagrange multiplier introduced to avoid overfitting to the logging policy \cite{joachims2018deep}. The optimization of Eq \eqref{eq:gen_ips} can be easily solved by gradient descent. When updating the classifier, we only use the discriminator's signals. Intuitively, even though annotations should contain the information about the true labels, the inverse is not necessary. The classifier is updated in a similar fashion,
\begin{align}
    \theta_C = \underset{\theta_C}{\text{argmin}}\frac{1}{NR}\sum_{n=1}^N\sum_{r=1}^R\frac{\big(\mathcal{L}_G(y_n^r)-\mu\big)G_{\theta_C}(y_n^r)}{G_0(y_n^r)}.
    \label{eq:clf_ips}
\end{align}
We follow the suggestions in \cite{joachims2018deep} to search the best $\mu$ in practice. 

\begin{figure}
\begin{subfigure}{0.235\textwidth}
  \centering
  \includegraphics[width=4.2cm]{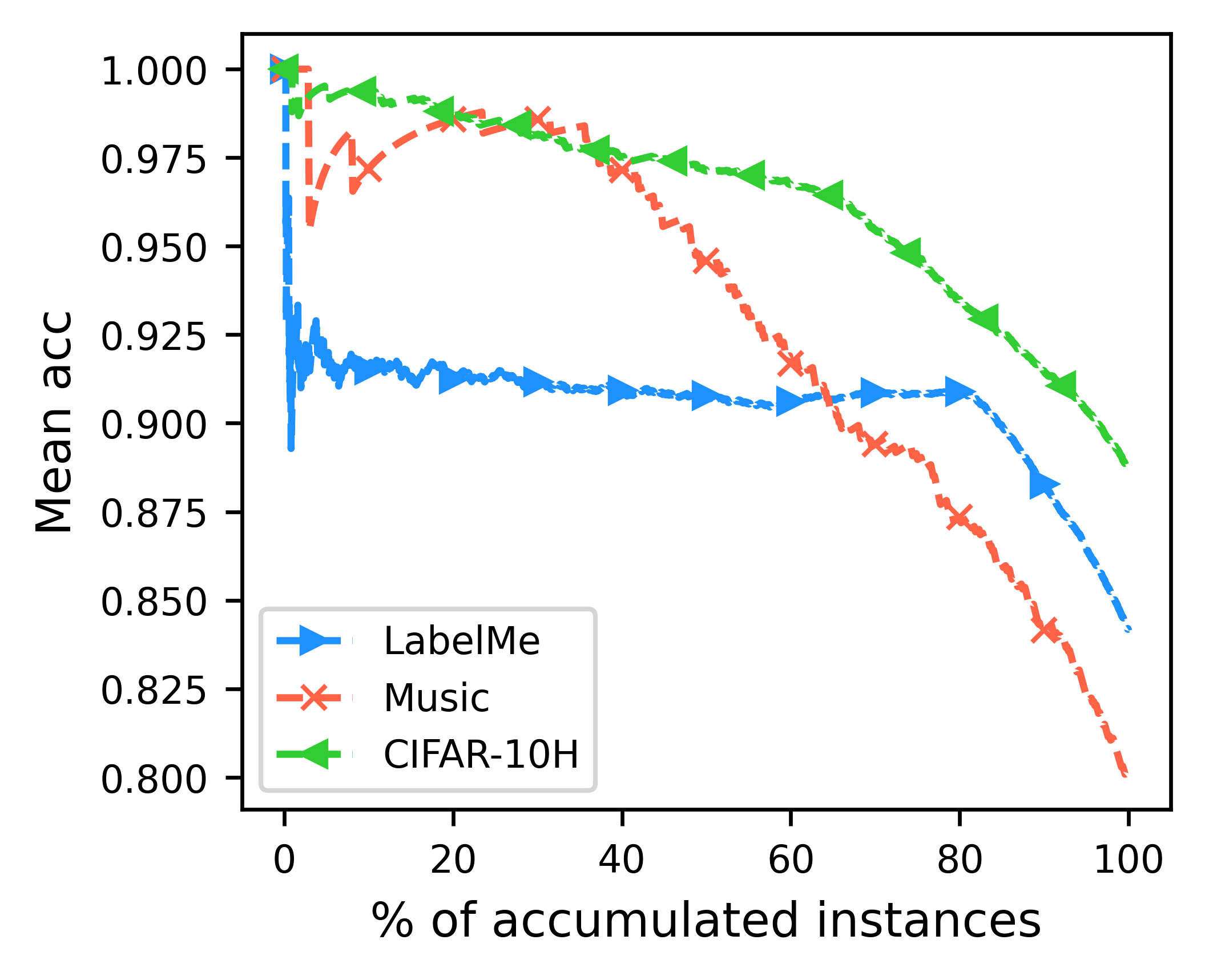}
  \caption{}
\end{subfigure}
\begin{subfigure}{0.235\textwidth}
  \centering
  \includegraphics[width=4.2cm]{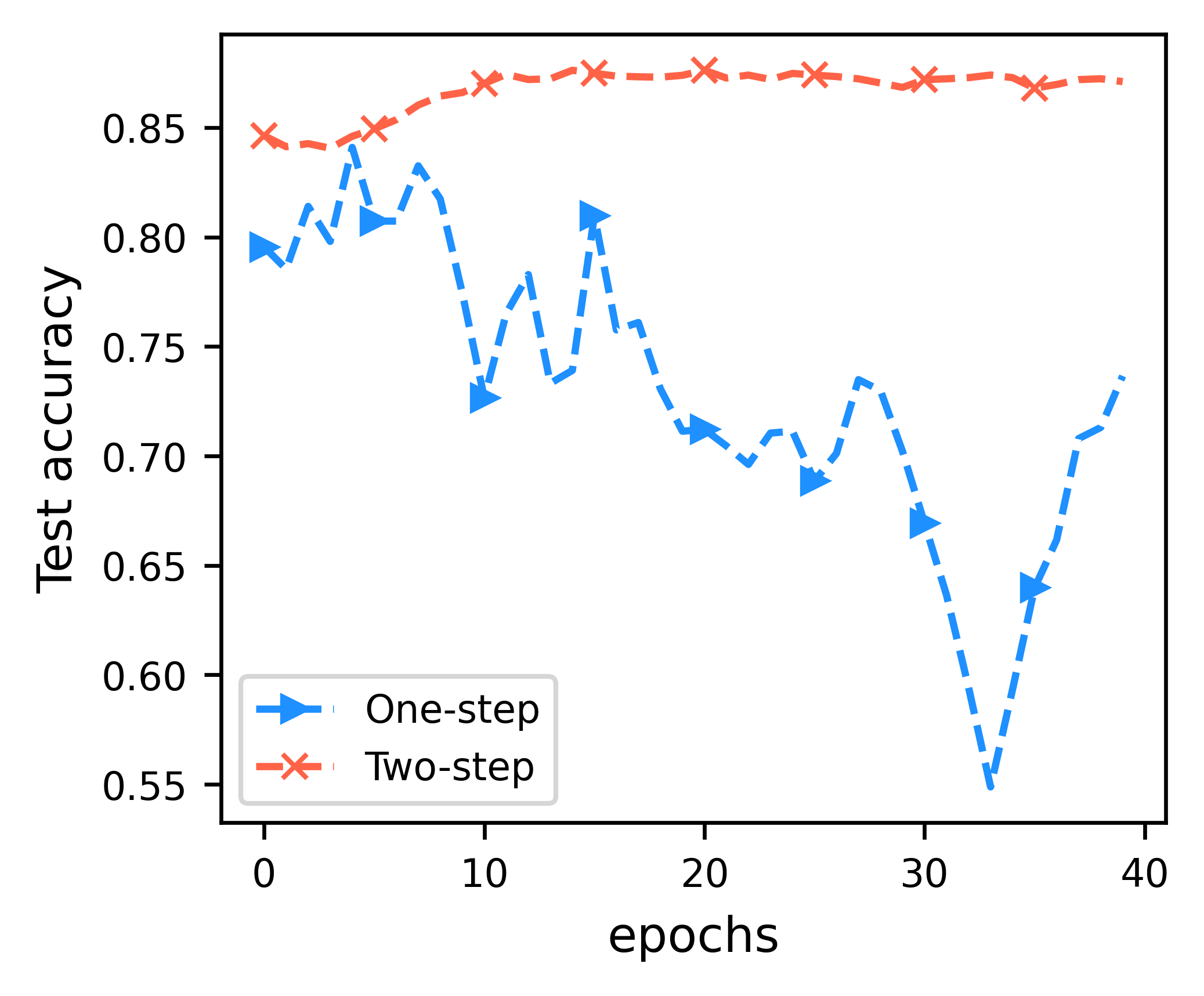}
  \caption{}
\end{subfigure}
\caption{Performance of two-step strategy. (a) Mean accuracy of accumulated instances with ascending order of entropy on three real-world datasets. (b) Comparison between one-step and two-step strategy on LabelMe dataset.}
\label{fig:two-step}
\vspace{-1em}
\end{figure}

\subsubsection{Two-step update for the generative module} 
\label{sec:two-step}
The generative process is controlled by the generator and the classifier. However, the coupling between the two components introduces difficulties in the estimation of them. For example, one component might overfit a particular pattern in the discriminative signal, and cause model collapse in the entire pipeline. In our empirical studies reported in Figure \ref{fig:two-step}(b), we observed test accuracy fluctuated a lot when we simply used gradients calculated by Eq \eqref{eq:gen_ips} and \eqref{eq:clf_ips} to update these two components together. Details of our experiment setup can be found in Section \ref{sec:exp}. 

Based on this finding, we adopt a two-step strategy to update the generator and the classifier alternatively. First, we found that
the principle behind our annotation selection also applied to our classifier: the entropy of the classifier's output strongly correlates with its accuracy. According to Figure \ref{fig:two-step}(a), the classifier obtains higher accuracy on instances with lower prediction entropy. Therefore, we decided to use the instances with low classification entropy to update the generator by Eq \eqref{eq:gen_ips}, as there the classifier's predictions are more likely to be accurate. Then, we use the updated generator on the rest of instances to update the classifier by Eq \eqref{eq:clf_ips}, where the classifier still has a high uncertainty to handle them.

A threshold $t$ is pre-selected to separate the instances; and we will discuss its influence on model training in Section \ref{exp:ablation}. Besides, to make the entire training process stable, we pre-train the classifier with the observed annotations using neural crowdsourcing algorithm proposed in \cite{rodrigues2018deep}, which is included as one of our baselines. With the initialized classifier, we also pre-train the generator and discriminator to provide good initialization of these components.  

%% file: experiments.tex
\begin{figure*}[!htp]
\begin{subfigure}{1\textwidth}
  \centering
  \includegraphics[width=15.6cm]{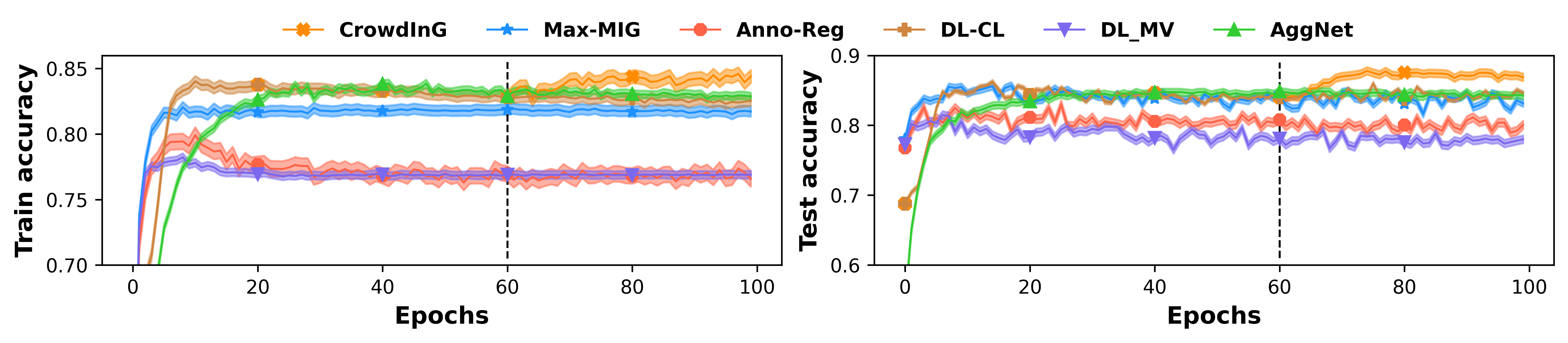}
  \caption{Results on LabelMe dataset.}
  \label{fig:sub3}
\end{subfigure}
\begin{subfigure}{1\textwidth}
  \centering
  \includegraphics[width=15.6cm]{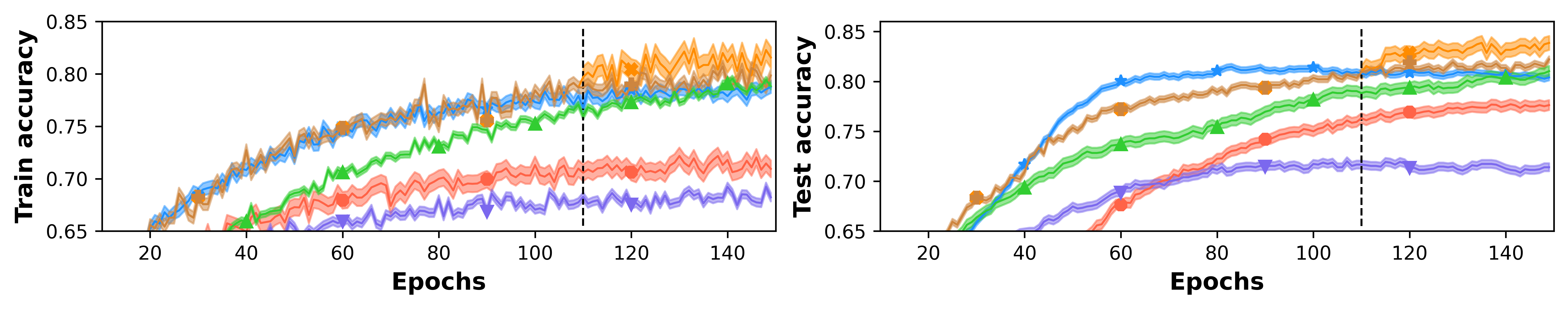}
  \caption{Results on Music dataset.}
  \label{fig:sub3}
\end{subfigure}
\begin{subfigure}{1\textwidth}
  \centering
  \includegraphics[width=15.6cm]{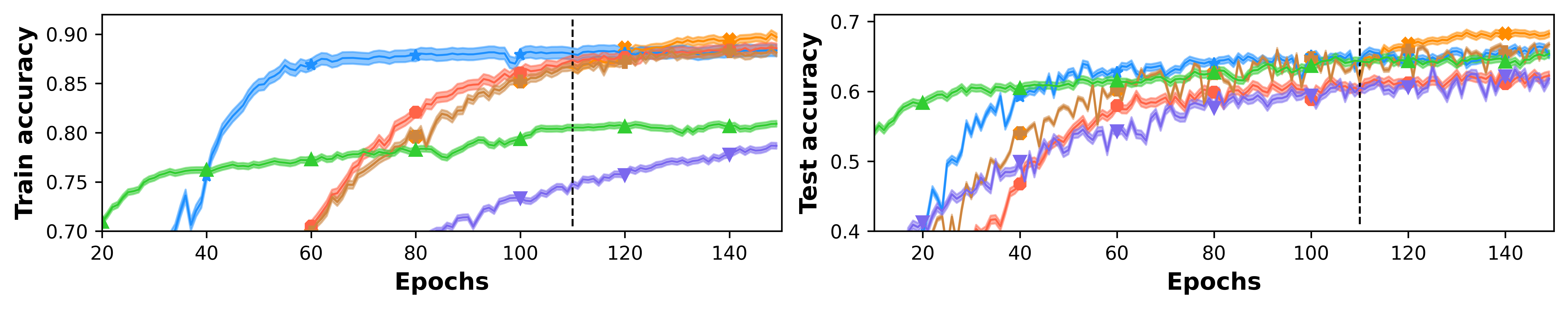}
  \caption{Results on CIFAR-10H dataset.}
  \label{fig:sub3}
\end{subfigure}
\caption{Results on three real-world datasets. Full CrowdInG training is applied after the dashed line.}
\label{fig:test_acc}
\vspace{-0.5em}
\end{figure*}

\section{Experiments}
\label{sec:exp}
In this section, we evaluate our proposed solution framework on three real-world datasets. The annotations were originally collected from Amazon Mechanical Turk (AMT) by the dataset creators. We compared with a rich set of state-of-the-art crowdsourcing algorithms that estimate the classifiers only with observed annotations. We are particularly interested in investigating \emph{how much human labor can be saved by our data augmentation solution}? We gradually removed an increasing number of annotations and compared with baselines. The result suggests significant annotation cost can be reduced with our generated annotations, while still maintaining the quality of the learnt classifier. Besides, since our model is the first effort to augment crowdsourced data for classifier training, we compared with models trained with annotations from other generative models for crowdsourced data. Finally, we performed extensive ablation analysis about our proposed model components and hyper-parameters to better understand the model's behavior.

\subsection{Datasets \& Implementation details}
\begin{figure}[!htp]
\centering
\begin{subfigure}{.24\textwidth}
  \centering
  \includegraphics[height=2.8cm]{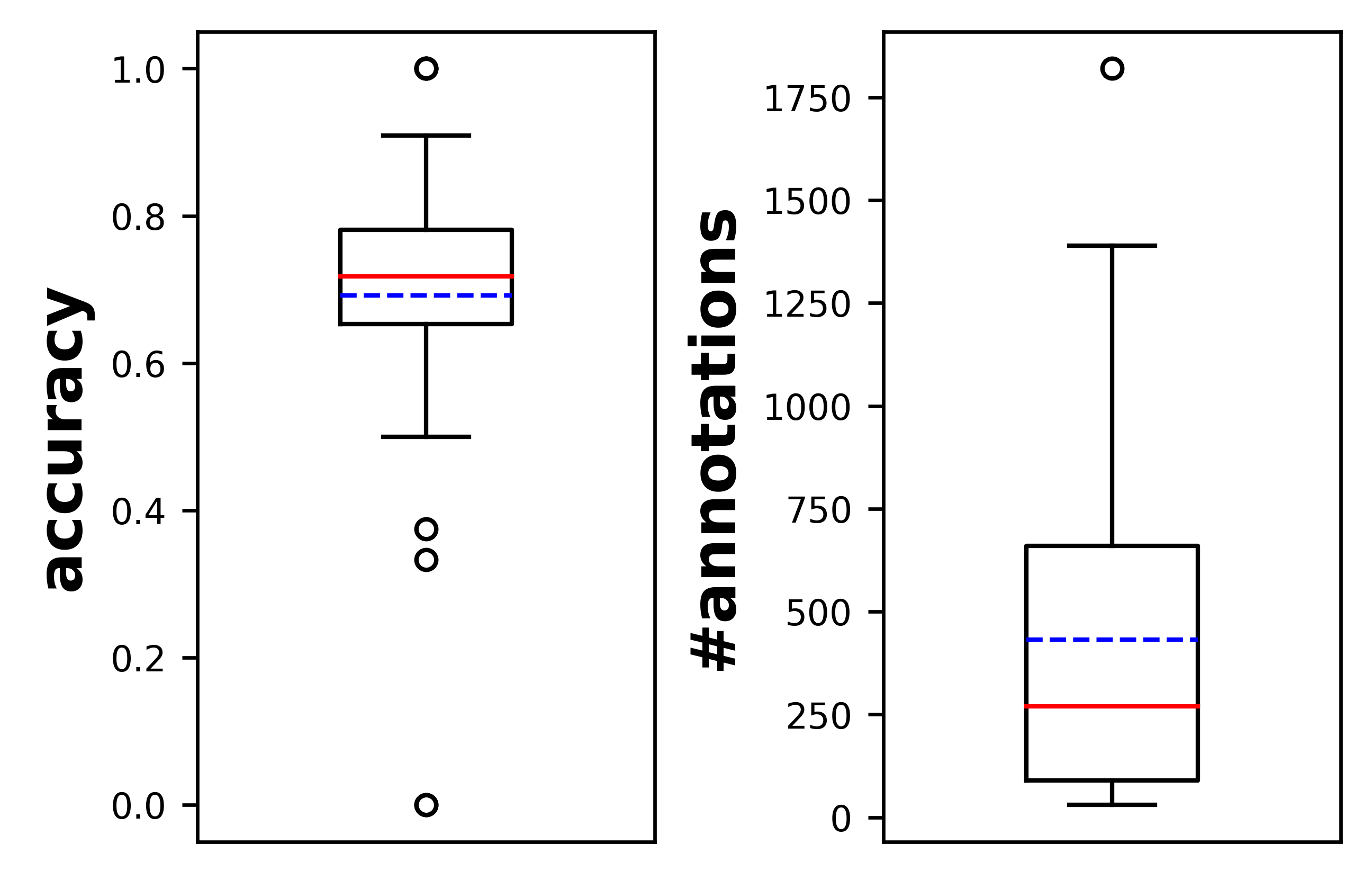}
  \caption{LabelMe}
  \label{fig:labelme}
\end{subfigure}%
\begin{subfigure}{.24\textwidth}
\centering
  \includegraphics[height=2.8cm]{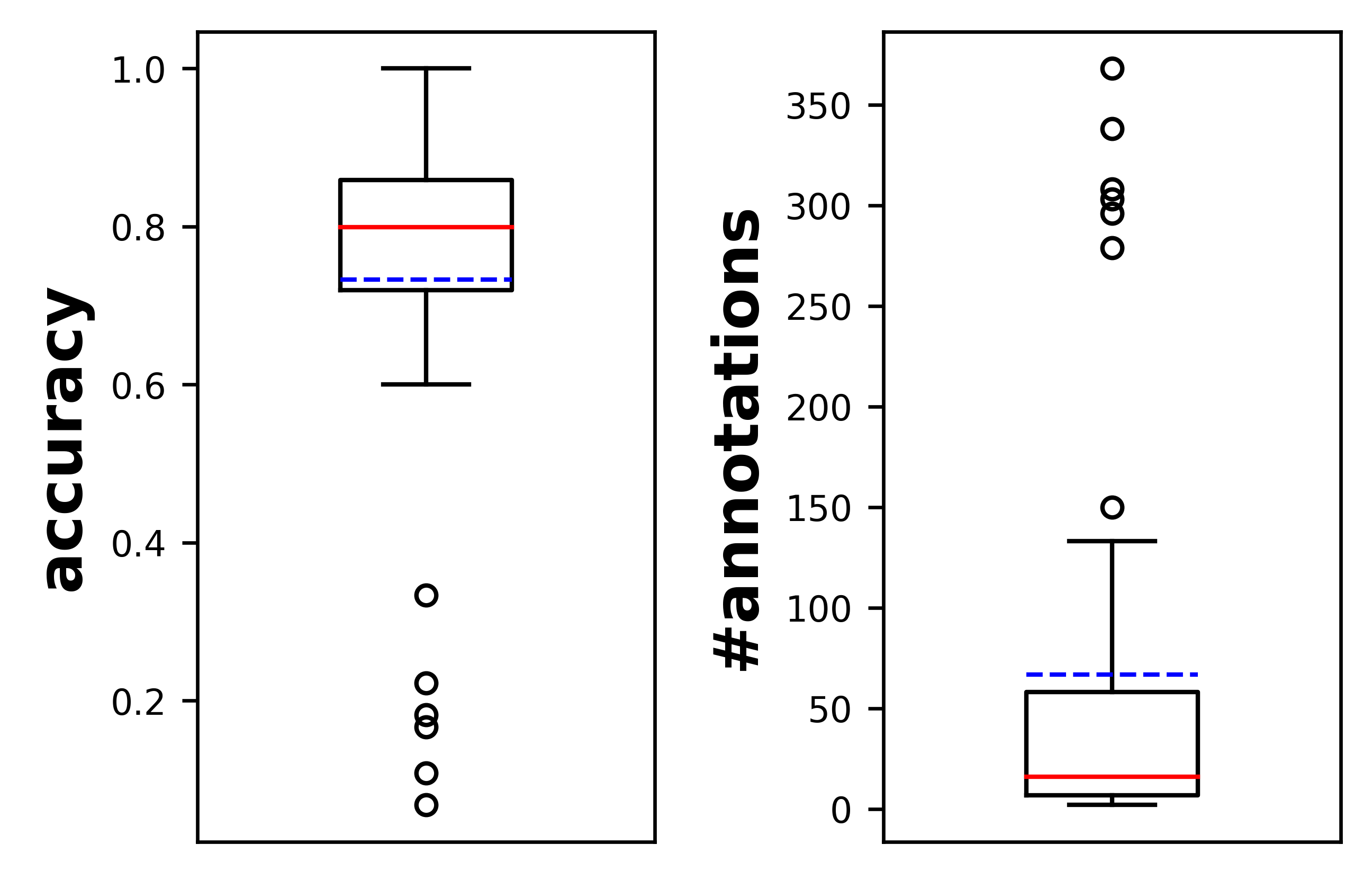}
  \caption{Music}
  \label{fig:music}
\end{subfigure}
\caption{Boxplots for the number of annotations and the accuracy of the AMT annotators for two real-world crowdsourcing datasets.}
\label{fig:annotator_analysis}
\vspace{-1em}
\end{figure}

We employed three real-world datasets for evaluations. \textbf{LabelMe} \cite{russell2008labelme, rodrigues2018deep} is an image classification dataset, which consists of 2,688 images from 8 classes, e.g., \emph{inside city}, \emph{street}, \emph{forest}, etc. 1,000 of them are labeled by 59 AMT annotators and the rest are used for validation and testing. Each image is labeled by 2.5 annotators on average. To enrich the training set, standard data augmentation techniques are applied on the training set, including horizontal flips, rescaling and shearing, following the setting in \cite{rodrigues2018deep}. We created 10,000 images for training eventually. \textbf{Music} \cite{rodrigues2014gaussian} is a music genre classification dataset, which consists of 1,000 samples of songs with 30 seconds in length from 10 classes, e.g., \emph{classical}, \emph{country}, \emph{jazz}, etc. 700 of them are labeled by 44 AMT annotators and the rest are left for testing. Each sample is labeled by 4.2 annotators on average. Figure \ref{fig:annotator_analysis} shows several important statistics of these two datasets. Specifically, we report the annotation accuracy and the number of annotations among the annotators. Both statistics vary considerably across annotators in these two datasets, which cause serious difficulties in classical crowdsourcing algorithms. \textbf{CIFAR-10H} \cite{peterson2019human} is another image classification dataset, which consists of 10,000 images from 10 classes, e.g., \emph{airplane}, \emph{bird}, \emph{cat}, etc., collected from the CIFAR-10 image dataset \cite{krizhevsky2009learning}. There were 2,571 annotators recruited and each annotator was asked to label 200 images. However, such large-scale annotations are typically expensive and rare in practice. To make this dataset closer to a realistic and challenging setting, we only selected a subset of low-quality annotators. The modified dataset has 8,687 images annotated by 103 AMT annotators. Each annotator still has 200 annotations with an average accuracy of 78.2\%; and each image has 2.37 annotations on average. The original 10,000 images validation set of CIFAR-10 is used as our testing set. 

\begin{figure*}[!htp]
\centering
  \centering
  \includegraphics[width=16.6cm]{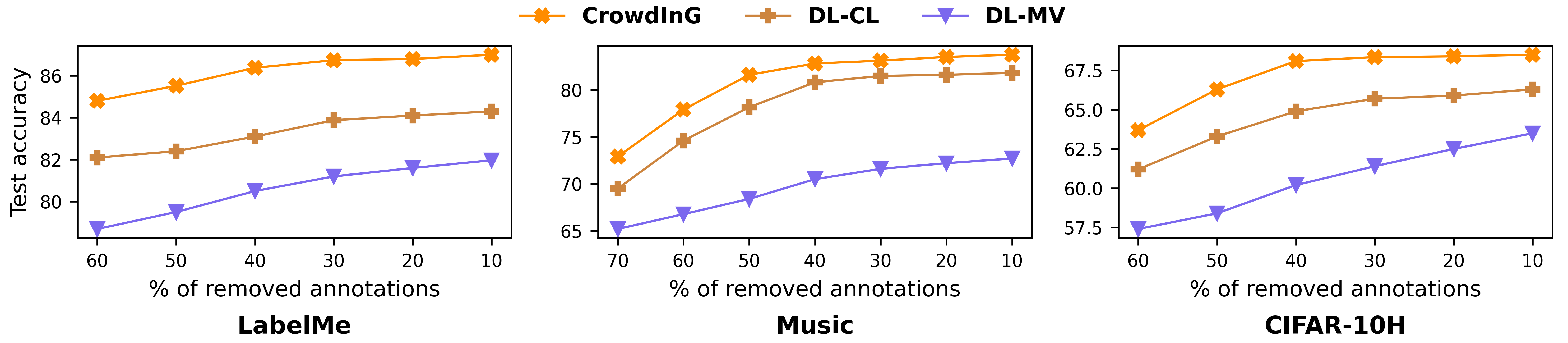}
  \label{fig:labelme}
\caption{Test accuracy with various proportion of removed annotations.}
\label{fig:remove}
\vspace{-0.5em}
\end{figure*}

To make the comparisons fair, all evaluated methods used the same classifier design (in both \model{} and baselines). On the LabelMe dataset, we adopted the same setting as in \cite{rodrigues2018deep}: we applied a pre-trained VGG-16 network followed by a fully connected (FC) layer with 128 units and ReLU activations, and a softmax output layer, using 50\% dropout. On the Music dataset, we also used a 128 units FC layer and softmax output layer. Batch normalization was performed in each layer. We disabled LCA on Music since there is no meaningful label correlation patterns. On the CIFAR-10H dataset, we used a VGG-16 network for the classifier. We used Adam optimizer with learning rates searched from $\{3.0 \times 10^{−4}, 2.0 \times 10^{−4}, 1.0 \times 10^{−4}, 1.0 \times 10^{−5}\}$ for both generative and discriminative modules. Scoring functions $g(\cdot)$ and $h(\cdot)$ are implemented by two-layer neural networks with 64 and 128 hidden units. In each epoch, we update the generative and discriminative modules for 5 times. With pre-training, we execute the training procedures for CrowdInG in the last 40 epochs. 
All experiments are repeated 5 times with different random seeds, and mean accuracy and standard derivation are reported. 

 \begin{table}[!htp]
\centering
\caption{Test accuracy of different augmentation methods.}
\label{tab:genereated_anno}
\begin{tabular}{| c @{\hspace{1\tabcolsep}} | c @{\hspace{1\tabcolsep}} | c @{\hspace{1\tabcolsep}} | c @{\hspace{1\tabcolsep}}|}
\hline
        & LabelMe & Music & CIFAR-10H \\ \hline
Doctor Net & 82.12{\small $\pm 0.43$} & 75.41{\small $\pm 0.42$}  & 67.23{\small $\pm 0.54$}\\
$\text{DL-CL}_{\text{+Self}}$ & 85.24{\small $\pm 0.51$}  & 82.56{\small $\pm 0.49$} & 64.94{\small $\pm 0.84$} \\ 
$\text{DL-CL}_{\text{+GCN}}$ & 82.74{\small $\pm 0.34$} & 81.42{\small $\pm 0.74$}  & 65.02{\small $\pm 0.61$} \\
$\text{DL-CL}_{\text{+GAN}}$ & 85.16{\small $\pm 0.26$} & 83.17{\small $\pm 0.48$} & 65.34{\small $\pm 0.32$} \\
$\text{DL-CL}_{\text{+InG}}$ & 85.42{\small $\pm 0.57$} & 83.38{\small $\pm 0.59$} & 66.17{\small $\pm 0.35$} \\
CrowdInG & \textbf{87.03}{\small $\pm 0.55$} &  \textbf{83.73}{\small $\pm 0.62$} & \textbf{68.85}{\small $\pm 0.47$} \\ \hline
\end{tabular}
\vspace{-0.5em}
\end{table}

\subsection{Classification performance}
\subsubsection{Baselines} We compared with a rich set of state-of-the-art baselines, which we briefly introduce here. \textbf{DL-MV}: annotations are first aggregated by majority vote, and then it trains a classifier based on the aggregated labels. \textbf{DL-CL} \cite{rodrigues2018deep}: a set of designated layers that capture annotators' confusions (the so-called Crowd Layer) are connected to the classifier, aiming to transform the predicted classifier's outputs to annotation distributions. \textbf{Anno-Reg} \cite{tanno2019learning}: trace regularization on confusion matrices is applied to improve the confusion estimation. \textbf{Max-MIG} \cite{cao2018maxmig}: a neural classifier and a label aggregation network are jointly trained using an information-theoretical loss function, correlated confusions among annotators are captured. \textbf{AggNet} \cite{albarqouni2016aggnet}: an EM-based deep model considering annotator sensitivity and specificity.

\subsubsection{Results \& analysis}
The classification accuracy of the learnt classifiers from different models on the three datasets are reported in Figure \ref{fig:test_acc}. Two things we should highlight: 1) as all models are learnt from crowdsourced data, the ground-truth labels on instances are \emph{unrevealed} to them in training. Therefore, a classifier's accuracy on training set is still a meaningful performance metric. 2) \model{} starts with the same classifier as obtained in DL-CL (as we used DL-CL to pre-train our classifier). On all datasets, we observe that even though DL-CL did not outperform the other baselines, after the training in CrowdInG starts, the classifier's performance got significantly improved. This proves the utility of our generated annotations for classifier training.
Besides, we also looked into the accuracy in individual classes and found by generating more annotations, \model{}'s performance on those easily confused classes got more improvement than the baselines. For example, for the class of \emph{open country} on LabelMe, the original annotation accuracy was only 51.5\%. DL-CL achieved 49.6\%  (i.e., the starting point of \model{}), and it was improved to 58.9\% after \model{} training. 
Compared with models that are designed for complex confusions, such as Max-MIG and AggNet, \model{} still outperformed them with a large margin. This indicates our generator has a stronger advantage in capturing complex confusions.

\subsection{Utility of augmented annotations}
\subsubsection{Experiment setup} We study the utility of augmented annotations from \model{}. On each dataset, we gradually removed an increasing number of observed annotations to investigate how different models' performance changes. We ensure that each instance has at least one annotation, such that we will only remove annotations rather than instances for classifier training. We compared with two representative baselines: 1) DL-MV, a typical majority-vote-based method, and 2) DL-CL, a typical DS-model-based method, to study their sensitivity on the sparsity of annotations. 

\subsubsection{Results \& analysis} 
We present the results in Figure \ref{fig:remove}. All models suffered from extreme sparsity when we removed a large portion of annotations (e.g., 60\%), but CrowdInG still enjoyed a consistent improvement against all baselines. DL-MV performed the worst, because with less redundant annotations, the quality of its aggregated labels deteriorated seriously.
When we looked into the detailed model update trace of \model{}, we found that the performance gain became larger after \model{} training. Again, because we used the classifier obtained from DL-CL as our starting point for \model{}, low-quality annotations were generated at the beginning of \model{} update. 
However, \model{} quickly improved once its discriminative module started to penalize those low-quality annotations.
The results strongly support that a great deal of human labor can be saved. On LabelMe and CIFAR-10H, CrowdInG performed closely to the baselines' best performance even with 60\% less annotations. Even on the most difficult dataset Music, about 10\% annotations can be saved by CrowdInG to achieve similar performance as DL-CL.

\subsection{Comparison with other augmentations}
\subsubsection{Baselines} As no existing method explicitly performs data augmentation for crowdsourced data, we consider several alternative data augmentation methods using various self-training or generative modeling techniques. Arguably, any generative model for crowdsourced data can be used for this purpose. 

In particular, we chose the following baselines. \textbf{Doctor Net} \cite{guan2018said}: each annotator is modeled by an individual neural network. When testing, annotations are predicted by annotator networks and then aggregated by weighted majority vote.  \textbf{$\text{DL-CL}_{\text{+Self}}$}: we complete the missing annotations using a pre-trained DL-CL model, and then train another DL-CL model based on the completed annotations. \textbf{$\text{DL-CL}_{\text{+GCN}}$}: we construct an annotator-instance bipartite graph based on the observed annotations, and fill in the missing links using a Graph Convolution Network (GCN) \cite{berg2017graph, kipf2016semi}. Then we train a DL-CL model using the expanded annotations. \textbf{$\text{DL-CL}_{\text{+GAN}}$}: we follow the same design in \cite{wang2017irgan}, which unifies generative and discriminative models into a GAN framework. We use DL-CL as the generative model. \textbf{$\text{DL-CL}_{\text{+InG}}$}: we directly train a DL-CL model on the expanded dataset provided by CrowdInG.

\subsubsection{Results \& analysis}
We present the test accuracy on all three datasets in Table \ref{tab:genereated_anno}. Doctor Net trains individual classifiers for each annotator, so that on datasets where annotations from each annotator are sufficient, such as CIFAR-10H, this model obtained satisfactory performance with the generated annotations. But on the other datasets where annotations are sparse in each annotator, its performance dropped a lot. In DL-CL type methods, the performance is generally improved. However, due to the simple class-dependent confusion assumption, such models' capacity to capture complex confusions is limited. As a result, even though GCN could capture more complex annotator-instance interactions, DL-CL still failed to benefit from it in $\text{DL-CL}_{\text{+GCN}}$. The added discriminator in $\text{DL-CL}_{\text{+GAN}}$ improved the performance; however, DL-CL still could not fully utilize the complex discriminative signals and failed to further improve the performance. $\text{DL-CL}_{\text{+InG}}$ performed better than the other baselines by directly using the annotations generated by \model{}, which suggests the  annotations generated under our criteria are generically helpful for other crowdsouring algorithms. 
 
 \begin{table}[!htp]
\centering
\caption{Test accuracy of different variants of \model{}}
\label{tab:ablation}
\begin{tabular}{|c @{\hspace{1\tabcolsep}} | c @{\hspace{1\tabcolsep}} | c @{\hspace{1\tabcolsep}} | c @{\hspace{1\tabcolsep}}|}
\hline
        & LabelMe & Music & CIFAR-10H \\ \hline
CrowdG  & 85.89{\small $\pm 0.47$} & 83.14{\small $\pm 0.28$} & 66.15{\small $\pm 0.34$} \\ 
$\text{CrowdInG}_{\text{U}}$ & 83.12{\small $\pm 0.39$} & 81.28{\small $\pm 0.51$}  & 67.12{\small $\pm 0.59$} \\ 
$\text{CrowdInG}_{\text{I}}$ & 84.34{\small $\pm 0.72$} & 82.24{\small $\pm 0.47$} & 66.90{\small $\pm 0.31$}\\ 
$\text{CrowdInG}_{\text{R}}$ & 86.17{\small $\pm 0.44$} & 82.74{\small $\pm 0.58$}  & 67.88{\small $\pm 0.62$} \\ 
CrowdInG & \textbf{87.03}{\small $\pm 0.55$} &  \textbf{83.73}{\small $\pm 0.62$} & \textbf{68.85}{\small $\pm 0.47$} \\  \hline
\end{tabular}
\vspace{-0.5em}
\end{table}
\subsection{Ablation study}
\label{exp:ablation}
\subsubsection{Analysis of different components in \model{}} To show the contributions of different components in \model{}, we varied the setting of our solution. We already showed the one-step training variant in Figure \ref{fig:two-step}, which suffered from serious model collapsing. To investigate the other components, we created the following variants. \textbf{CrowdG}: the information loss defined in Eq \eqref{eq:info_lb} is removed. \textbf{$\text{CrowdInG}_{\text{U}}$}: the generator only considers classifier's outputs, annotator features and random noise, but not the instance features. \textbf{$\text{CrowdInG}_{\text{I}}$}: the generator only considers classifier's outputs, instance features and random noise, but not the annotator features. 
\textbf{$\text{CrowdInG}_{\text{R}}$}: the annotation selection is kept, but instead we randomly select an equal number of generated annotations as the authentic ones for discriminator update.

We reported the test accuracy on three datasets in Table \ref{tab:ablation}. By maximizing the mutual information, CrowdInG outperformed CrowdG with a considerable margin. We further investigated the generated annotations and found the annotations generated by CrowdG were more random, which could not be easily linked to the classifier's output. $\text{CrowdInG}_{\text{U}}$ performed poorly when the number of annotations per annotator was limited, such as on LabelMe and Music datasets, but worked better when annotations per annotator are adequate, such as on CIFAR-10H. This again proves more annotations are needed to better model annotators' confusions. $\text{CrowdInG}_{\text{I}}$ performed better because by taking instance features, the generator can model more complicated confusions with respect to instance features. 
$\text{CrowdInG}_{\text{R}}$ bypassed the data imbalance issue; but without focusing on difficult annotations, it still cannot fully unleash the potential of generated annotations.

\begin{figure}[!htp]
\centering
  \centering
  \includegraphics[width=8.1cm]{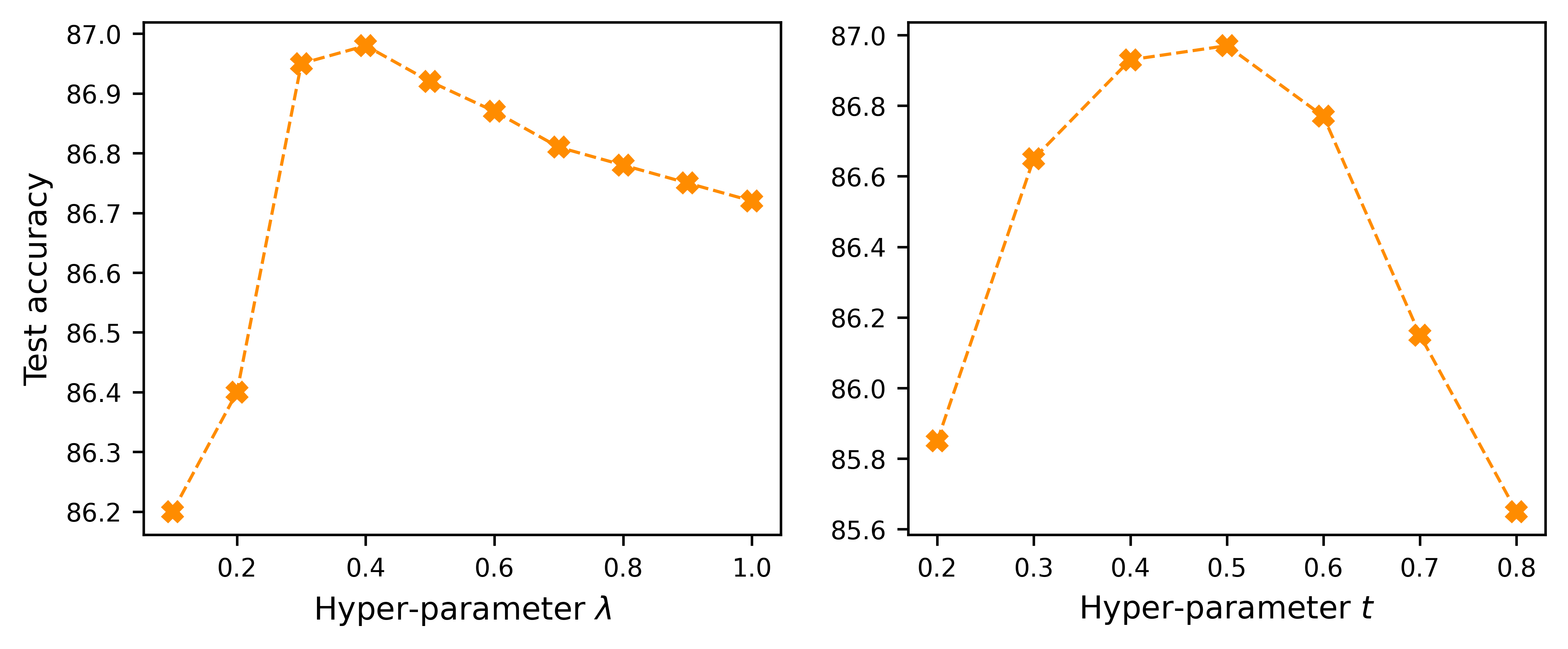}
  \label{fig:labelme}
\caption{Performance under different hyper-parameter settings on LabelMe dataset.}
\label{fig:hyper}
\vspace{-1em}
\end{figure}

\subsubsection{Hyper-parameter analysis.} We studied the sensitivity of hyper-parameters $\lambda$ and $t$ in CrowdInG. Specifically, $\lambda$ controls the degree of the information regularization in Eq \eqref{eq_obj}, we varied it from 0.1 to 1. $t$ controls the grouping of instances used for classifier update; and we varied it from 0.2 to 0.8. Due to space limit, we only report the results on LabelMe, similar observations were also  obtained on the other two datasets.

The model's performance  under different hyper-parameter settings is illustrated in Figure \ref{fig:hyper}. We can clearly observe that the performance is boosted when appropriate hyper-parameters are chosen. Small $\lambda$ poses weak information regularization to the generator, and thus the generated annotations are less informative for classifier training. Large $\lambda$ slightly hurts the performance because strong regularization weakens the ability of the generator to capture complex confusions related to instance and annotator features. We can observe similar trend on $t$. To avoid model collapse, a moderate $t$ is needed to restrict the classifier training, but a large $t$ will hurt the performance more. Because with a large $t$, very few instances will be selected for classifier training, so that the classifier can hardly be updated.